\documentclass{article}

\usepackage[accepted]{icml-like} 

\usepackage{times}

\usepackage{graphicx} 
\usepackage{subfigure} 

\usepackage{booktabs}

\usepackage{natbib}

\usepackage{algorithm}
\usepackage{algorithmic}
\usepackage{afterpage}

\usepackage{hyperref}
\usepackage{url}

\usepackage{amsmath}

\begin{document} 

\twocolumn[
\icmltitle{Binarized Neural Networks: Training Neural Networks with Weights and Activations Constrained to $+1$ or $-1$}

\icmlauthor{Matthieu Courbariaux*$^1$}{matthieu.courbariaux@gmail.com}
\icmlauthor{Itay Hubara*$^2$}{itayhubara@gmail.com}
\icmlauthor{Daniel Soudry$^3$}{daniel.soudry@gmail.com}
\icmlauthor{Ran El-Yaniv$^2$}{rani@cs.technion.ac.il}
\icmlauthor{Yoshua Bengio$^{1,4}$}{yoshua.umontreal@gmail.com}
\icmladdress{
$^1$Universit\'{e} de Montr\'{e}al \\
$^2$Technion - Israel Institute of Technology \\
$^3$Columbia University \\
$^4$CIFAR Senior Fellow \\
*Indicates equal contribution. Ordering determined by coin flip.
}

\icmlkeywords{deep learning, neural network, convolution, MLP, ConvNet, running, training, 
    MNIST, CIFAR-10, SVHN, binary, low precision, memory, hardware, GPU, FPGA, acceleration, speed-up, low power}

\vskip 0.3in
]

\begin{abstract}

We introduce a method to train Binarized Neural Networks (BNNs) - neural networks with binary weights and activations at run-time. At training-time the binary weights and activations are used for computing the parameters gradients. During the forward pass, BNNs drastically reduce memory size and accesses, and replace most arithmetic operations with bit-wise operations, which is expected to  substantially improve power-efficiency.
To validate the effectiveness of BNNs we conduct two sets of experiments on the Torch7 and Theano frameworks. On both, BNNs achieved nearly state-of-the-art results over the MNIST, CIFAR-10 and SVHN datasets.
Last but not least, we wrote a binary matrix multiplication GPU kernel with which it is possible to run our MNIST BNN 7 times faster 
than with an unoptimized GPU kernel, without suffering any loss in classification accuracy. The code for training and running our BNNs is available on-line.

\end{abstract}

\section*{Introduction}

Deep Neural Networks (DNNs) have substantially pushed Artificial Intelligence (AI) limits in a wide range of tasks,
including but not limited to object recognition from images~\citep{Krizhevsky-2012-small,Szegedy-et-al-arxiv2014},
speech recognition~\citep{Hinton-et-al-2012,Sainath-et-al-ICASSP2013},
statistical machine translation~\citep{Devlin-et-al-ACL2014,Sutskever-et-al-NIPS2014,Bahdanau-et-al-ICLR2015-small},
Atari and Go games \citep{Mnih-et-al-2015,Silver-et-al-2016},
and even abstract art \citep{Mordvintsev-et-al-2015}.

Today, DNNs are almost exclusively trained on one or many very fast and power-hungry Graphic Processing Units (GPUs) \citep{Coates-et-al-2013}.
As a result, it is often a challenge to run DNNs on target low-power devices,
and substantial research efforts are invested in speeding up DNNs at run-time on both general-purpose \citep{Vanhoucke-et-al-2011,Gong-et-al-2014,Romero-et-al-2014,Han-et-al-2015}
and specialized computer hardware \citep{Farabet-et-al-2011-a, Farabet-et-al-2011-b, Pham-et-al-2012, Chen-et-al-ACM2014, Chen-et-al-IEEE2014, Esser-et-al-2015}.

This paper makes the following contributions:
\begin{itemize}
    
    \item We introduce a method to train Binarized-Neural-Networks (BNNs), neural networks with binary weights and activations, at run-time,
        and when computing the parameters gradients at train-time (see Section \ref{sec:BNN}).
    
    \item We conduct two sets of experiments, each implemented on a different framework, namely Torch7 \citep{Torch-2011} and Theano \citep{bergstra+al:2010-scipy,Bastien-Theano-2012},
        which show that it is possible to train BNNs on MNIST, CIFAR-10 and SVHN and achieve nearly state-of-the-art results (see Section \ref{sec:benchmark}).
    
    \item  We show that during the forward pass (both at run-time and train-time), BNNs drastically reduce memory consumption (size and number of accesses), 
        and replace most arithmetic operations with bit-wise operations,
        which potentially lead to a substantial increase in power-efficiency (see Section \ref{sec:power}).
        Moreover, a binarized CNN can lead to binary convolution kernel repetitions; We argue that dedicated hardware could reduce the time complexity by $60\%$ .
        
    \item Last but not least, we programed a binary matrix multiplication GPU kernel with which it is possible to run 
        our MNIST BNN 7 times faster than with an unoptimized GPU kernel, 
        without suffering any loss in classification accuracy (see Section \ref{sec:faster}). 
    
    \item The code for training and running our BNNs is available on-line (In both Theano framework \footnote{\url{https://github.com/MatthieuCourbariaux/BinaryNet}}
     and Torch framework
    \footnote{\url{https://github.com/itayhubara/BinaryNet}}).
  
\end{itemize}

\section{Binarized Neural Networks}
\label{sec:BNN}

In this section, we detail our binarization function, show how we use it to compute the parameters gradients,
and how we backpropagate through it.

\subsection{Deterministic vs Stochastic Binarization}

When training a BNN, we constrain both the weights and the activations to either $+1$ or $-1$.
Those two values are very advantageous from a hardware perspective, as we explain in Section \ref{sec:faster}.
In order to transform the real-valued variables into those two values, 
we use two different binarization functions, as in \citep{Courbariaux-et-al-2015}.
Our first binarization function is deterministic:
\begin{equation}
    x^b = {\rm Sign}(x) = \left\{ \begin{array}{ll}
                        +1 & \mbox{if $x \geq 0$},\\
                        -1 & \mbox{otherwise},\end{array} \right.
\end{equation}
where $x^b$ is the binarized variable (weight or activation) and $x$ the real-valued variable.
It is very straightforward to implement and works quite well in practice.
Our second binarization function is stochastic:
\begin{align}
\label{eq:sampled-wb}
    x^b = \left\{ \begin{array}{ll}
            +1 & \mbox{with probability $p = \sigma(x)$},\\
            -1 & \mbox{with probability $1-p$},\end{array} \right. 
\end{align}
where $\sigma$ is the {\em ``hard sigmoid''} function:
\begin{equation}
    \sigma(x) = {\rm clip}(\frac{x+1}{2},0,1) = \max(0,\min(1,\frac{x+1}{2})).
\end{equation}
The stochastic binarization is more appealing than the sign function, 
but harder to implement as it requires the hardware to generate random bits when quantizing.
As a result, we mostly use the deterministic binarization function (i.e, the sign function),
with the exception of {\em activations at train-time} in some of our experiments.

\subsection{Gradient Computation and Accumulation}

Although our BNN training method uses binary weights and activation to compute the parameters gradients, the real-valued gradients of the weights are accumulated in real-valued variables, as per Algorithm \ref{alg:train}.
Real-valued weights are likely required for Stochasic Gradient Descent (SGD) to work at all.
SGD explores the space of parameters in small and noisy steps, and that noise is
{\em averaged out} by the stochastic gradient contributions  accumulated in each weight.
Therefore, it is important to keep sufficient resolution for these accumulators, which at first glance
suggests that high precision is absolutely required.

Moreover, adding noise to weights and activations when {\em computing} the parameters gradients
provide a form of regularization that can help to generalize better, 
as previously shown with variational weight noise~\citep{Graves-2011-practical},
Dropout~\citep{Srivastava-master-small,Srivastava14} and DropConnect~\citep{Wan+al-ICML2013-small}.
Our method of training BNNs can be seen as a variant of Dropout, 
in which instead of randomly setting half of the activations to zero when computing the parameters gradients,
we binarize both the activations and the weights.

\begin{algorithm}
\begin{algorithmic}
    \REQUIRE a minibatch of inputs and targets $(a_0,a^*)$,
    previous weights $W$, previous BatchNorm parameters $\theta$, 
    weights initialization coefficients from \citep{GlorotAISTATS2010-small} $\gamma$,
    and previous learning rate $\eta$.
    \ENSURE updated weights $W^{t+1}$, updated BatchNorm parameters $\theta^{t+1}$ and updated learning rate $\eta^{t+1}$.
    
    \STATE \COMMENT{1. Computing the parameters gradients:}
    
    \STATE \COMMENT{1.1. Forward propagation:}
    \FOR{$k=1$ to $L$}  
        \STATE $W_k^b \leftarrow {\rm Binarize}(W_k)$
        \STATE $s_k \leftarrow a_{k-1}^b W_k^b$
        \STATE $a_k \leftarrow {\rm BatchNorm}(s_k, \theta_k)$
        \IF{$k < L$}
            \STATE $a_k^b \leftarrow {\rm Binarize}(a_k)$
        \ENDIF
    \ENDFOR
    
    \STATE \COMMENT{1.2. Backward propagation:}
    \STATE \COMMENT{Please note that the gradients are not binary.}
    \STATE Compute $g_{a_L}=\frac{\partial C}{\partial a_L}$ knowing $a_L$ and $a^*$
    \FOR{$k=L$ to $1$} 
        \IF{$k < L$}
            \STATE $g_{a_k} \leftarrow g_{a_k^b} \circ 1_{|a_k|\leq 1}$
        \ENDIF
        \STATE $(g_{s_k}, g_{\theta_k}) \leftarrow {\rm BackBatchNorm}(g_{a_k}, s_k,\theta_k)$
        
        \STATE $g_{a_{k-1}^b} \leftarrow g_{s_k} W_k^{b}$
        \STATE $g_{W_k^b} \leftarrow g_{s_k}^{\top} a_{k-1}^b$

    \ENDFOR
    
    \STATE \COMMENT{2. Accumulating the parameters gradients:}
    \FOR{$k=1$ to $L$}
        \STATE $\theta_k^{t+1} \leftarrow {\rm Update}(\theta_k, \eta, g_{\theta_k})$
        \STATE $W_k^{t+1} \leftarrow {\rm Clip}({\rm Update}(W_k, \gamma_k \eta, g_{W_k^b}),-1,1)$
        \STATE $\eta^{t+1} \leftarrow \lambda \eta$
    \ENDFOR
    
\end{algorithmic}
\caption{
Training a BNN. $C$ is the cost function for minibatch, $\lambda$ - the learning rate decay factor and $L$ the number of layers.
$\circ$ indicates element-wise multiplication.
The function Binarize() specifies how to (stochastically or deterministically) binarize the activations and weights,  
and Clip(), how to clip the weights.
BatchNorm() specifies how to batch-normalize the activations,
using either batch normalization \citep{Ioffe+Szegedy-2015} or its shift-based variant we describe in Algorithm \ref{alg:BN}.
BackBatchNorm() specifies how to backpropagate through the normalization.
Update() specifies how to update the parameters when their gradients are known,
using either ADAM \citep{kingma2014adam} or the shift-based AdaMax we describe in Algorithm \ref{alg:adamax}.
}
\label{alg:train}
\end{algorithm}

\begin{algorithm} [h]
\begin{algorithmic}
    \REQUIRE Values of $x$ over a mini-batch: $B=\{x_{1\ldots m}\}$; 
        Parameters to be learned: $\gamma$, $\beta$
    \ENSURE $\{y_i = {\rm BN}({x_i,}{\gamma,\beta})\}$
    \STATE $\mu_B \leftarrow \frac{1}{m}\sum_{i=1}^m x_i$ \COMMENT{mini-batch mean}
    \STATE $C(x_i) \leftarrow (x_i-\mu_B)$
    \COMMENT{centered input}
    \STATE \!$\sigma_B^2\! \leftarrow \!\! \frac{1}{m}\sum_{i=1}^m\! (C(x_i)\!\!\ll\gg\!\! AP2(C(x_i)))\!$ \COMMENT{apx variance}
    \STATE $\hat{x_i} \leftarrow C(x_i)\ll\gg AP2((\sqrt{\sigma_B^2+\epsilon})^{-1})$ \COMMENT{normalize}
    \STATE $y_i \leftarrow AP2(\gamma)\ll\gg\hat{x_i}$ \COMMENT{scale and shift} 
\end{algorithmic}
\caption{Shift based Batch Normalizing Transform, applied to activation x over a mini-batch. $AP2(x) = \mathrm{sign}(x) \times 2^{\mathrm{round}(\mathrm{log2}{|x|})}$
    is the approximate power-of-2 \protect\footnotemark, and  $\ll\gg$ stands for \textbf{both }left and right binary shift.}
\label{alg:BN}
\end{algorithm}
\afterpage{
\footnotetext{Hardware implementation of AP2 is as simple as extracting the index of the most significant bit from the number's binary representation.}
}


\begin{algorithm}[h]
\begin{algorithmic}
    \REQUIRE Values of $x$ over a mini-batch: $B=\{x_{1\ldots m}\}$; 
        Parameters to be learned: $\gamma$, $\beta$
    \ENSURE $\{y_i = {\rm BN}({x_i,}{\gamma,\beta})\}$
    \STATE $\mu_B \leftarrow \frac{1}{m}\sum_{i=1}^m x_i$ \COMMENT{mini-batch mean}
    \STATE $C(x_i) \leftarrow (x_i-\mu_B)$
    \COMMENT{centered input}
    \STATE \!$\sigma_B^2\! \leftarrow \!\! \frac{1}{m}\sum_{i=1}^m\! (C(x_i)\!\!\ll\gg\!\! AP2(C(x_i)))\!$ \COMMENT{apx variance}
    \STATE $\hat{x_i} \leftarrow C(x_i)\ll\gg AP2((\sqrt{\sigma_B^2+\epsilon})^{-1})$ \COMMENT{normalize}
    \STATE $y_i \leftarrow AP2(\gamma)\ll\gg\hat{x_i}$ \COMMENT{scale and shift}
\end{algorithmic}
\caption{Shift based Batch Normalizing Transform, applied to activation $(x)$ over a mini-batch. Where AP2 is the approximate power-of-2 and  $\ll\gg$ stands for \textbf{both }left and right binary shift.}
\label{alg:BN}
\end{algorithm}

\begin{algorithm}[h]
\begin{algorithmic}
    \REQUIRE Previous parameters $\theta_{t-1}$ and their gradient $g_t$, and learning rate $\alpha$.
    \ENSURE Updated parameters $\theta_t$
    \STATE \COMMENT{Biased 1st and 2nd raw moment estimates:}
    \STATE $m_t \gets \beta_1 \cdot m_{t-1} + (1-\beta_1) \cdot g_t$  
    \STATE $v_t \gets \max (\beta_2 \cdot v_{t-1} , |g_t| )$ 
    \STATE \COMMENT{Updated parameters:}
    \STATE $\theta_t \gets \theta_{t-1} - (\alpha \ll\gg (1-\beta_1)) \cdot \hat{m}\ll\gg v_t^{-1})$
\end{algorithmic}
\caption{Shift based AdaMax learning rule \citep{kingma2014adam}.
$g^2_t$ indicates the element-wise square $g_t \circ g_t$. 
Good default settings are $\alpha=2^{-10}$, $1-\beta_1=2^{-3}$, $1-\beta_2=2^{-10}$ . 
All operations on vectors are element-wise. 
With $\beta_1^t$ and $\beta_2^t$ we denote $\beta_1$ and $\beta_2$ to the power $t$.}
\label{alg:adamax}
\end{algorithm}

\begin{algorithm}[h]
\begin{algorithmic}
    \REQUIRE a vector of 8-bit inputs $a_0$, 
    the binary weights $W^b$, and the BatchNorm parameters $\theta$.
    \ENSURE the MLP output $a_L$.
    
    \STATE \COMMENT{1. First layer:}
    \STATE $a_1 \leftarrow 0$
    \FOR{$n=1$ to $8$}  
        \STATE $a_1 \leftarrow a_1 + 2^{n-1} \times {\rm XnorDotProduct(a_0^n,W^b_1)}$
    \ENDFOR
    \STATE $a_1^b \leftarrow {\rm Sign(BatchNorm}(a_1,\theta_1))$
    
    \STATE \COMMENT{2. Remaining hidden layers:}
    \FOR{$k=2$ to $L-1$}  
        \STATE $a_k \leftarrow {\rm XnorDotProduct}(a_{k-1}^b,W^b_k)$
        \STATE $a_k^b \leftarrow {\rm Sign(BatchNorm}(a_k,\theta_k))$
    \ENDFOR
    
    \STATE \COMMENT{3. Output layer:}
    \STATE $a_L \leftarrow {\rm XnorDotProduct}(a_{L-1}^b,W^b_L)$
    \STATE $a_L \leftarrow {\rm BatchNorm}(a_L,\theta_L)$
    
\end{algorithmic}
\caption{Running a BNN. $L$ is the number of layers.}
\label{alg:run}
\end{algorithm}

\subsection{Propagating Gradients Through Discretization}

The derivative of the sign function is zero almost everywhere, making it apparently
incompatible with backpropagation, since the exact gradient of the cost with respect to
the quantities before the discretization (pre-activations or weights) would be zero.
Note that this remains true even if stochastic quantization is used.
\citet{Bengio-arxiv2013} studied the question of estimating or propagating
gradients through stochastic discrete neurons. They found
in their experiments that the fastest training was obtained when using
the ``straight-through estimator,'' previously introduced in \citet{Hinton-Coursera2012}'s lectures.

We follow a similar approach but use the version of the straight-through
estimator that takes into account the saturation effect, and does use
deterministic rather than stochastic sampling of the bit.
Consider the sign function quantization
\[
q = {\rm Sign}(r),
\]
and assume that an estimator $g_q$ of the gradient $\frac{\partial C}{\partial q}$
has been obtained (with the straight-through estimator when needed).
Then, our straight-through estimator of $\frac{\partial C}{\partial r}$ is simply
\begin{equation}
  \label{eq:straight-through-gradient}
g_r = g_q 1_{|r|\leq 1}.
\end{equation}
Note that this preserves the gradient's information and cancels the
gradient when $r$ is too large. 
Not cancelling the gradient when $r$ is too large significantly worsens the performance. 
The use of this straight-through estimator is illustrated in Algorithm \ref{alg:train}.
The derivative $1_{|r|\leq 1}$ can
also be seen as propagating the gradient through {\em hard tanh}, which
is the following piece-wise linear activation function:
\begin{equation}
    {\rm Htanh}(x) = {\rm Clip}(x,-1,1) = \max(-1,\min(1,x)).
\end{equation}

For hidden units, we use the sign function non-linearity to obtain binary
activations, and for weights we combine two ingredients:
\begin{itemize}
\item Constrain each real-valued weight between -1 and 1, by projecting $w^r$
  to -1 or 1 when the weight update brings $w^r$ outside of $[-1,1]$, i.e., clipping the weights
  during training, as per Algorithm \ref{alg:train}. 
  The real-valued weights would otherwise grow very large without any impact on the binary weights.
  \item When using a weight $w^r$, quantize it using $w^b = {\rm Sign}(w^r)$.
\end{itemize}
This is consistent with the gradient canceling when $|w^r|>1$, according
to Eq.~\ref{eq:straight-through-gradient}.



\subsection{Shift based Batch Normalization} 

Batch Normalization (BN) \citep{Ioffe+Szegedy-2015}, accelerates the training and also seems to reduces the overall impact of the weights' scale. 
The normalization noise may also help to regularize the model. 
However, at train-time, BN requires many multiplications (calculating the standard deviation and dividing by it), 
namely, dividing by the running variance (the weighted mean of the training set activation variance). 
Although the number of scaling calculations is the same as the number of neurons, in the case of ConvNets this number is quite large. 
For example, in the CIFAR-10 dataset (using our architecture), the first convolution layer, consisting of only $128\times3\times3$ filter masks, 
converts an image of size $3\times32\times32$ to size $3\times128\times28\times28$, 
which is two orders of magnitude larger than the number of weights. To achieve the results that BN would obtain, 
we use a shift-based batch normalization (SBN) technique. detailed in Algorithm \ref{alg:BN}. 
SBN approximates BN almost without multiplications. 
In the experiment we conducted we did not observe accuracy loss when using the shift based BN algorithm instead of the vanilla BN algorithm.
 
\subsection{Shift based AdaMax} 
 
The ADAM learning rule \citep{kingma2014adam} also seems to reduce the impact of the weight scale. 
Since ADAM requires many multiplications, 
we suggest using instead the shift-based AdaMax we detail in Algorithm \ref{alg:adamax}. 
In the experiment we conducted we did not observe accuracy loss when using the shift-based AdaMax algorithm instead of the vanilla ADAM algorithm.

 

\subsection{First Layer}

In a BNN, only the binarized values of the weights and activations are used in all calculations.
As the output of one layer is the input of the next, all the layers inputs are binary, with the exception of the first layer.
However, we do not believe this to be a major issue.
First, in computer vision, the input representation typically has much fewer channels (e.g, Red, Green and Blue) 
than internal representations (e.g, 512).
As a result, the first layer of a ConvNet is often the smallest convolution layer, 
both in terms of parameters and computations \citep{Szegedy-et-al-arxiv2014}.

Second, it is relatively easy to handle continuous-valued inputs as fixed point numbers,
with $m$ bits of precision. For example, in the common case of $8$-bit fixed point inputs:
\begin{align}
    s & = x \cdot w^b \\
    s & = \sum_{n=1}^{8} {2^{n-1} (x^n \cdot w^b),}
\end{align}
where $x$ is a vector of 1024 8-bit inputs, $x_1^8$ is the most significant bit of the first input, $w^b$ is a vector of 1024 1-bit weights,
and $s$ is the resulting weighted sum.
This trick is used in Algorithm \ref{alg:run}.

\section{Benchmark Results}
\label{sec:benchmark}

\begin{table*}[t]
\protect\caption{Classification test error rates of DNNs trained on MNIST (MLP architecture without unsupervised pretraining), CIFAR-10 (without data augmentation) and SVHN.}
\centering{}%
\scalebox{.75}{
\begin{tabular}{lcccr}
\hline 
Data set  & MNIST & SVHN & CIFAR-10 & \tabularnewline
\hline 
\hline 
\multicolumn{5}{c}{Binarized activations+weights, during training and test}\tabularnewline
\hline 
\hline 
BNN (Torch7) & 1.40\% & 2.53\% & 10.15\% & \tabularnewline
BNN (Theano) & 0.96\% & 2.80\% & 11.40\% & \tabularnewline
Committee Machines' Array \citep{Baldassi2015} & 1.35\% & - & - \tabularnewline
\hline 
\hline 
\multicolumn{5}{c}{Binarized weights, during training and test}\tabularnewline
\hline 
\hline 
BinaryConnect \citep{Courbariaux-et-al-2015} & 1.29$\pm$ 0.08\% & 2.30\% & 9.90\% & \tabularnewline
\hline 
\hline 
\multicolumn{5}{c}{Binarized activations+weights, during test}\tabularnewline
\hline 
\hline 
EBP \citep{Cheng-et-al-2015} & 2.2$\pm$ 0.1\% & - & - & \tabularnewline
Bitwise DNNs \citep{Kim-et-al-2016} & 1.33\% & - & - & \tabularnewline
\hline 
\hline 
\multicolumn{5}{c}{Ternary weights, binary activations, during test}\tabularnewline
\hline 
\hline 
\citep{hwang-et-al-2014} & 1.45\% & - & - & \tabularnewline

\hline 
\hline 
\multicolumn{5}{c}{No binarization (standard results)}\tabularnewline
\hline 
\hline 
Maxout Networks \citep{Goodfellow2013a} & 0.94\% & 2.47\% & 11.68\% & \tabularnewline
Network in Network \citep{Lin} & - & 2.35\% & 10.41\% & \tabularnewline
Gated pooling \citep{lee-et-al-2015}& - & 1.69\% & 7.62\% & \tabularnewline
\hline 
\end{tabular}\label{tab:benchmarks}
}
\end{table*}

\begin{figure}[ht]
\caption{
Training curves of a ConvNet on CIFAR-10 depending on the method.
The dotted lines represent the training costs (square hinge losses)
and the continuous lines the corresponding validation error rates.
Although BNNs are slower to train, they are nearly as accurate as 32-bit float DNNs.
}
\begin{center}
\centerline{\includegraphics[width=.47\textwidth]{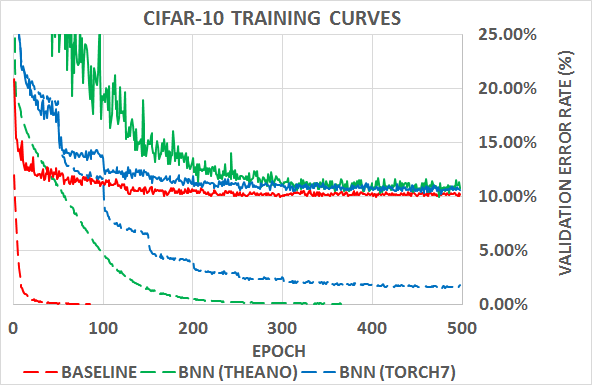}}
\end{center}
\label{fig:curves}
\end{figure}

\begin{figure}[ht]
\caption{Binary weight filters, sampled from of the first convolution layer.
Since we have only $2^{k^{2}}$ unique 2D filters (where $k$ is the
filter size), filter replication is very common.
For instance, on our CIFAR-10 ConvNet,
only 42\% of the filters are unique. }
\begin{centering}
\includegraphics[viewport=140bp 220bp 480bp 550bp,clip,width=0.5\columnwidth]{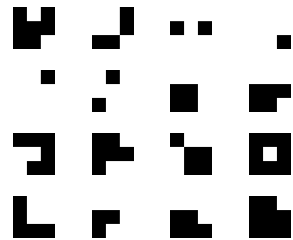}
\par\end{centering}
\label{Fig:filters}
\end{figure}

We conduct two sets of experiments, each based on a different framework, namely Torch7 \citep{Torch-2011} and Theano \citep{bergstra+al:2010-scipy,Bastien-Theano-2012}. 
Other than the framework, the two sets of experiments are very similar:
\begin{itemize}
    \item In both sets of experiments, we obtain near state-of-the-art results with BNNs on MNIST, CIFAR-10 and the SVHN benchmark datasets.
    \item In our Torch7 experiments, the activations are {\em stochastically} binarized at train-time,
        whereas in our Theano experiments they are {\em deterministically} binarized.
    \item In our Torch7 experiments, we use the {\em shift-based BN and AdaMax} variants, 
        which are detailed in Algorithms \ref{alg:BN} and \ref{alg:adamax},
        whereas in our Theano experiments, we use {\em vanilla BN and ADAM}.
\end{itemize} 

\subsection{MLP on MNIST (Theano)} 

MNIST is an image classification benchmark dataset \citep{LeCun+98}.
It consists of a training set of 60K and a test set of 10K 28 $\times$ 28 gray-scale images representing digits ranging from 0 to 9.
In order for this benchmark to remain a challenge, we did not use any convolution, data-augmentation, preprocessing or unsupervised learning.
The MLP we train on MNIST consists of 3 hidden layers of 4096 binary units (see Section \ref{sec:BNN})
and a L2-SVM output layer; L2-SVM has been shown to perform 
better than Softmax on several classification benchmarks \citep{Tang-wkshp-2013,Lee-et-al-2014}.
We regularize the model with Dropout \citep{Srivastava-master-small,Srivastava14}.
The square hinge loss is minimized with the ADAM adaptive
learning rate method \citep{kingma2014adam}.
We use an exponentially decaying global learning rate, as per Algorithm \ref{alg:train},
and also scale the learning rates of the weights with their initialization coefficients from \citep{GlorotAISTATS2010-small}, 
as suggested by \citet{Courbariaux-et-al-2015}.
We use Batch Normalization with a minibatch of size 100 to speed up the training.
As is typical, we use the last 10K samples of the training set as a validation set for early stopping and model selection.
We report the test error rate associated with the best validation error rate after 1000 epochs
(we do not retrain on the validation set).
The results are reported in Table \ref{tab:benchmarks}.

\subsection{MLP on MNIST (Torch7)} 

We use a similar architecture as in our Theano experiments, without dropout, and with 2048 binary units per layer instead of 4096. 
Additionally, we use the shift base AdaMax and BN (with a minibatch of size 100)  instead of the vanilla implementations, to reduce the number of multiplications. 
Likewise, we decay the learning rate by using a 1-bit right shift every 10 epochs. 
The results are presented in Table \ref{tab:benchmarks}.

\subsection{ConvNet on CIFAR-10 (Theano)}

CIFAR-10 is an image classification benchmark dataset.
It consists of a training set of size 50K and a test set of size 10K, where instance are 32 $\times$ 32 color images representing 
airplanes, automobiles, birds, cats, deer, dogs, frogs, horses, ships and trucks.
We do not use any preprocessing or data-augmentation (which can really be a game changer for this dataset \citep{Graham-2014}).
The architecture of our ConvNet is the same architecture as \citet{Courbariaux2015}'s except for the binarization of the activations.
\citet{Courbariaux-et-al-2015}'s architecture is itself mainly inspired by VGG \citep{Simonyan2015}.
The square hinge loss is minimized with ADAM.
We use an exponentially decaying learning rate, as we did for MNIST.
We scale the learning rates of the weights with their initialization coefficients from \citep{GlorotAISTATS2010-small}.
We use Batch Normalization with a minibatch of size 50 to speed up the training.
We use the last 5000 samples of the training set as a validation set.
We report the test error rate associated with the best validation error rate after 500 training epochs
(we do not retrain on the validation set).
The results are presented in Table \ref{tab:benchmarks} and Figure \ref{fig:curves}.

\subsection{ConvNet on CIFAR-10 (Torch7)} 

We use the same architecture as in our Theano experiments. 
We apply shift-based AdaMax and BN (with a minibatch of size 200) instead of the vanilla implementations to reduce the number of multiplications. 
Likewise, we decay the learning rate by using a 1-bit right shift every 50 epochs. 
The results are presented in Table \ref{tab:benchmarks} and Figure \ref{fig:curves}.

\subsection{ConvNet on SVHN}

SVHN is also an image classification benchmark dataset.
It consists of a training set of size 604K examples and a test set of size 26K, where instances are 32 $\times$ 32 color images representing digits ranging from 0 to 9.
In both sets of experiments, we follow the same procedure used for the CIFAR-10 experiments, 
with a few notable exceptions: we use half the number of units in the convolution layers, and we train for 200 epochs instead of 500 
(because SVHN is a much larger dataset than CIFAR-10).
The results are given in Table \ref{tab:benchmarks}.

\section{Very Power Efficient in Forward Pass}
\label{sec:power}

\begin{table}[t]
\protect\caption{Energy consumption of multiply-accumulations  \citep{Horowitz2014} }
\label{sample-table-1-1}
\centering{}%
\scalebox{.75}{
\begin{tabular}{lcc}
\hline 
Operation & MUL  & ADD \tabularnewline
\hline 
 8bit Integer & 0.2pJ & 0.03pJ\tabularnewline
32bit Integer  & 3.1pJ & 0.1pJ\tabularnewline
16bit Floating Point & 1.1pJ & 0.4pJ\tabularnewline
 32tbit Floating Point  & 3.7pJ & 0.9pJ\tabularnewline
\hline 
\end{tabular}\label{Tb:Add_MUL_Horowitz}
}
\end{table}

\begin{table}[t]
\protect\caption{Energy consumption of memory accesses \citep{Horowitz2014} }
\label{sample-table-1-1-1}
\centering{}%
\scalebox{.75}{
\begin{tabular}{lc}
\hline 
Memory size  & 64-bit memory access \tabularnewline
\hline 
 8K & 10pJ\tabularnewline
32K & 20pJ\tabularnewline
 1M & 100pJ\tabularnewline
 DRAM & 1.3-2.6nJ\tabularnewline
\hline 
\end{tabular}\label{TB:Memory}
}
\end{table}

Computer hardware, be it general-purpose or specialized, is composed of memories, arithmetic operators and control logic.
During the forward pass (both at run-time and train-time), BNNs drastically reduce memory size and accesses, 
and replace most arithmetic operations with bit-wise operations,
which might lead to a great increase in power-efficiency.
Moreover, a binarized CNN can lead to binary convolution kernel repetitions, and we argue that dedicated hardware could reduce the time complexity by $60\%$ .

\subsection{Memory Size and Accesses}

Improving computing performance has always been and remains a challenge.
Over the last decade, power has been the main constraint on performance \citep{Horowitz2014}.
This is why much research effort has been devoted to reducing the energy consumption of neural networks. 
\citet{Horowitz2014} provides rough numbers for the computations' energy consumption (the given numbers are for 45nm technology)
as summarized in Tables \ref{Tb:Add_MUL_Horowitz} and \ref{TB:Memory}.
Importantly, we can see that memory accesses typically consume more energy than arithmetic operations,
and {\em memory access' cost augments with memory size}.
In comparison with 32-bit DNNs, BNNs require 32 times smaller memory size {\em and} 32 times fewer memory accesses. This is expected to reduce energy consumption drastically (i.e., more than 32 times).

\subsection{XNOR-Count}

Applying a DNN mainly consists of convolutions and matrix multiplications.
The key arithmetic operation of deep learning is thus the multiply-accumulate operation. 
Artificial neurons are basically multiply-accumulators computing weighted sums of their inputs.
In BNNs, both the activations and the weights are constrained to either $-1$ or $+1$.
As a result, most of the 32-bit floating point multiply-accumulations are replaced by 1-bit XNOR-count operations.
This could have a big impact on deep learning dedicated hardware.
For instance, a 32-bit floating point multiplier costs about 200 Xilinx FPGA slices \citep{Govindu-et-al-2004,Beauchamp-et-al-2006},
whereas a 1-bit XNOR gate only costs a single slice.

\subsection{Exploiting Filter Repetitions}

When using a ConvNet architecture with binary weights, the number of unique
filters is bounded by the filter size. For example, in our implementation
we use filters of size $3\times3$, so the maximum number of unique
2D filters is $2^{9}=512$. However, this should not prevent expanding
the number of feature maps beyond this number, since the actual filter
is a 3D matrix. Assuming we have $M_{\ell}$ filters in the $\ell$
convolutional layer, we have to store a 4D weight matrix of size $M_{\ell}\times M_{\ell-1}\times k\times k$.
Consequently, the number of unique filters is $2^{k^{2}M_{\ell-1}}$.
When necessary, we apply each filter on the map and perform the required
multiply-accumulate (MAC) operations (in our case, using XNOR and popcount operations).
Since we now have binary filters, many 2D filters of size $k\times k$
repeat themselves. By using dedicated hardware/software, we can apply
only the unique 2D filters on each feature map and sum the result
wisely to receive each 3D filter's convolutional result. Note that an
inverse filter (i.e., {[}-1,1,-1{]} is the inverse of {[}1,-1,1{]})
can also be treated as a repetition; it is merely a multiplication
of the original filter by -1. For example, in our ConvNet architecture
trained on the CIFAR-10 benchmark, there are only 42\% unique filters
per layer on average. Hence we can reduce the number of the XNOR-popcount
operations by 3.

\section{Seven Times Faster on GPU at Run-Time}
\label{sec:faster}

\begin{figure}[h]
\caption{
The first three columns represent the time it takes to perform a $8192\times8192\times8192$ (binary) matrix multiplication
on a GTX750 Nvidia GPU, depending on which kernel is used. We can see that our XNOR kernel 
is 23 times faster than our baseline kernel and 3.4 times faster than cuBLAS.
The next three columns represent the time it takes to run the MLP from Section \ref{sec:benchmark} on the full MNIST test set.
As MNIST's images are not binary, the first layer's computations are always performed by the baseline kernel.
The last three columns show that the MLP accuracy does not depend on which kernel is used.
}
\begin{center}
\centerline{\includegraphics[width=.47\textwidth]{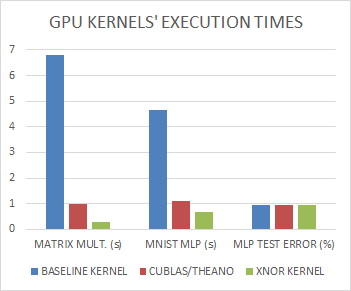}}
\end{center}
\label{fig:kernels}
\end{figure}

It is possible to speed up GPU implementations of BNNs,
by using a method sometimes called SIMD (single instruction, multiple data) within a register (SWAR). 
The basic idea of SWAR is to {\em concatenate} groups of 32 binary variables into 32-bit registers,
and thus obtain a 32-times speed-up on bitwise operations (e.g, XNOR).
Using SWAR, it is possible to evaluate 32 connections with only 3 instructions:
\begin{align}
    \label{eq:popc}
    a_1 += {\rm popcount(xnor}(a_0^{32b},w^{32b}_1)),
\end{align}
where $a_1$ is the resulting weighted sum, and $a_0^{32b}$ and $w^{32b}_1$ are the concatenated inputs and weights.
Those 3 instructions (accumulation, popcount, xnor) take $1+4+1=6$ clock cycles on recent Nvidia GPUs 
(and if they were to become a fused instruction, it would only take a single clock cycle).
Consequently, we obtain a theoretical Nvidia GPU speed-up of factor of $32/6 \approx 5.3$.
In practice, this speed-up is quite easy to obtain
as the memory bandwidth to computation ratio is also increased by 6 times.

In order to validate those theoretical results, we programed two GPU kernels:
\begin{itemize}
    \item The first kernel (baseline) is a quite unoptimized matrix multiplication kernel. 
    \item The second kernel (XNOR) is nearly identical to the baseline kernel, 
        except that it uses the SWAR method, as in Equation (\ref{eq:popc}).
\end{itemize}
The two GPU kernels return identical outputs when their inputs are constrained to $-1$ or $+1$ (but not otherwise).
The XNOR kernel is about {\em 23 times faster than the baseline kernel} and {\em 3.4 times faster than cuBLAS}, 
as shown in Figure \ref{fig:kernels}.
Last but not least, the MLP from Section \ref{sec:benchmark} runs 7 times faster with the XNOR kernel
than with the baseline kernel, without suffering any loss in classification accuracy 
(see Figure \ref{fig:kernels}).

\section{Discussion and Related Work} 

Until recently, the use of extremely low-precision networks (binary in the extreme case) was believed to be highly destructive to the network performance \citep{courbariaux+al-TR2014}. \citet{Soudry-et-al-NIPS2014-small,Cheng2015} showed the contrary by showing that good performance could be achieved even if all neurons and weights are binarized to $\pm 1$ . This was done using Expectation BackPropagation (EBP), a variational Bayesian approach, which infers networks with binary weights and neurons by updating the posterior distributions over the weights. These distributions are updated by differentiating their parameters (e.g., mean values) via the back propagation (BP) algorithm. \citet{Esser-et-al-2015} implemented a fully binary network at run time using a very similar approach to EBP, showing significant improvement in energy efficiency. The drawback of EBP is that the binarized parameters were only used during inference. 

The probabilistic idea behind EBP was extended in the BinaryConnect algorithm of \citet{Courbariaux-et-al-2015}. In BinaryConnect, the real-valued version of the weights is saved and used as a key reference for the binarization process. The binarization noise is independent between different weights, either by construction (by using stochastic quantization) or by assumption (a common simplification; see Spang (1962). The noise would have little effect on the next neuron's input because the input is a summation over many weighted neurons. Thus, the real-valued version could be updated by the back propagated error by simply ignoring the binarization noise in the update. Using this method, \citet{Courbariaux-et-al-2015} were the first to binarize weights in CNNs and achieved near state-of-the-art performance on several datasets. They also argued that noisy weights provide a form of regularization, which could help to improve generalization, as previously shown in \citep{Wan+al-ICML2013-small}. This method binarized weights while still maintaining full precision neurons.

\citet{Lin-et-al-2015} carried over the work of \citet{Courbariaux-et-al-2015} to the back-propagation process by quantizing the representations at each layer of the network, to convert some of the remaining multiplications into binary shifts by restricting the neurons values of power-of-two integers. \citet{Lin-et-al-2015}'s work and ours seem to share similar characteristics . However, their approach continues to use full precision weights during the test phase. Moreover, \citet{Lin-et-al-2015} quantize the neurons only during the back propagation process, and not during forward propagation. 

Other research \cite{Baldassi2015} showed that fully binary training and testing is possible in an array of committee machines with randomized input, where only one weight layer is being adjusted. \citet{Ounded2016} and \citet{Gong2015} aimed to compress a fully trained high precision network by using a quantization or matrix factorization methods. These methods required training the network with full precision weights and neurons, thus requiring numerous MAC operations avoided by the proposed BNN algorithm. \citet{hwang-et-al-2014} focused on a fixed-point neural network design and achieved performance almost identical to that of the floating-point architecture. \citet{kim-et-al-2014} provided evidence that DNNs with ternary weights, used on a dedicated circuit, consume very low power and can be operated with only on-chip memory, at run time. \citet{Sung2016} also indicated satisfactory empirical performance of neural networks with 8-bit precision. \citet{Kim2015} \emph{retrained} neural networks with binary weights and activations.

So far, to the best of our knowledge, no work has succeeded in binarizing weights \emph{and} neurons, at the inference phase \emph{and} the entire training phase of a deep network. This was achieved in the present work. We relied on the idea that binarization can be done stochastically, or be approximated as random noise. This was previously done for the weights by \citet{Courbariaux-et-al-2015}, but our BNNs extend this to the activations. Note that the binary activations are especially important for ConvNets, where there are typically many more neurons than free weights. This allows highly efficient operation of the binarized DNN at run time, and at the forward propagation phase during training. Moreover, our training method has almost no multiplications, and therefore might be implemented efficiently in dedicated hardware. However, we have to save the value of the full precision weights. This is a remaining computational bottleneck during training, since it requires relatively high energy resources. Novel memory devices might be used to alleviate this issue in the future; see e.g. \cite{Soudry}.

\section*{Conclusion}
 
We have introduced BNNs, DNNs with binary weights and activations at run-time and when computing the parameters gradients at train-time (see Section \ref{sec:BNN}).
We have conducted two sets of experiments on two different frameworks, Torch7 and Theano,
which show that it is possible to train BNNs on MNIST, CIFAR-10 and SVHN, and achieve nearly state-of-the-art results (see Section \ref{sec:benchmark}).
Moreover, during the forward pass (both at run-time and train-time), 
BNNs drastically reduce memory size and accesses, and replace most arithmetic operations with bit-wise operations,
which might lead to a great increase in power-efficiency (see Section \ref{sec:power}).
Last but not least, we programed a binary matrix multiplication GPU kernel with which it is possible to run our MNIST MLP 7 times faster 
than with an unoptimized GPU kernel, without suffering any loss in classification accuracy (see Section \ref{sec:faster}).
Future works should explore how to extend the speed-up to train-time (e.g., by binarizing some gradients),
and also extend benchmark results to other models (e.g, RNN) and datasets (e.g, ImageNet).
 
\section*{Acknowledgments}

We would like to express our appreciation to Elad Hoffer, for his technical assistance and constructive comments. We thank our fellow MILA lab members who took the time to read the article and give us some feedback. 
We thank the developers of Torch, \cite{Torch-2011} a Lua based environment, and Theano \citep{bergstra+al:2010-scipy,Bastien-Theano-2012}, a Python library which allowed us to easily develop a fast and optimized code for GPU. 
We also thank the developers of Pylearn2 \citep{pylearn2_arxiv_2013} and Lasagne \citep{dieleman-et-al-2015}, 
two Deep Learning libraries built on the top of Theano. 
We thank Yuxin Wu for helping us compare our GPU kernels with cuBLAS.
We are also grateful for funding from CIFAR, NSERC, IBM, Samsung, and the Israel Science Foundation (ISF).

{\small
\bibliography{strings,strings-shorter,aigaion,ml,binary_net,BinaryNets}
\bibliographystyle{icml-like-no-url}
}
\end{document}